\documentclass[runningheads,orivec]{llncs}
\usepackage[T1]{fontenc}
\usepackage{graphicx}

\usepackage{iciap}
\usepackage{iciapabbrv}

\usepackage{algorithm}
\usepackage{algorithmicx}
\usepackage{subcaption}
\usepackage{algpseudocode}
\usepackage{booktabs}       
\usepackage{amsmath}        
\usepackage{amssymb}        
\usepackage{xcolor}         
\usepackage{microtype}      
\usepackage{multirow}       
\usepackage{orcidlink}
\usepackage{hyperref}
\graphicspath{{./images/}}

\title{\methname: A multi-expert architecture for continual anomaly detection}
\author{Malihe Dahmardeh\inst{1}\orcidlink{0009-0006-5564-3355} \and Francesco Setti\inst{1,2}\orcidlink{0000-0002-0015-5534}}
\authorrunning{M. Dahmardeh et al.}
\institute{Department of Engineering for Innovation Medicine, University of Verona, Italy \and Qualyco S.r.l., strada le Grazie 15, Verona, Italy\\
Corresponding Author: \email{malihe.dahmardeh@univr.it}
}

\usepackage[dvipsnames]{xcolor}

\usepackage{xspace}
\newcommand{\methname}{MECAD\xspace}

\begin{document}

\maketitle

\begin{abstract}
In this paper we propose \methname, a novel approach for continual anomaly detection using a multi-expert architecture. Our system dynamically assigns experts to object classes based on feature similarity and employs efficient memory management to preserve the knowledge of previously seen classes. 
By leveraging an optimized coreset selection and a specialized replay buffer mechanism, we enable incremental learning without requiring full model retraining. Our experimental evaluation on the MVTec AD dataset demonstrates that the optimal 5-expert configuration achieves an average AUROC of 0.8259 across 15 diverse object categories while significantly reducing knowledge degradation compared to single-expert approaches.
This framework balances computational efficiency, specialized knowledge retention, and adaptability, making it well-suited for industrial environments with evolving product types.

\end{abstract}

\section{Introduction}


Anomaly Detection (AD) has become increasingly relevant in industrial applications, where detecting defects and irregularities in manufacturing processes is critical for quality control and operational efficiency. Traditional AD methods, particularly those based on supervised learning, require extensive labeled data, which can be expensive and time-consuming to collect. As a result, unsupervised AD approaches have gained traction, allowing systems to identify anomalies without relying on explicitly labeled abnormal samples.

One of the primary limitations of existing AD models is that they are trained on specific object categories and cannot generalize well when deployed in dynamic environments where new objects and variations frequently emerge. In industrial production lines, for example, manufacturing conditions, product types, and defect patterns change rapidly, necessitating models that can adapt to new object classes efficiently without requiring retraining from scratch. This adaptability is crucial for ensuring continued performance without incurring the high cost of dataset re-annotation and model re-optimization.

Continual learning (CL) provides a promising approach to address these challenges. By enabling models to learn sequentially from new tasks while preserving knowledge from previously learned tasks, CL mitigates catastrophic forgetting and allows for long-term adaptation. However, applying CL to anomaly detection presents unique difficulties, as conventional CL techniques are typically designed for classification problems rather than unsupervised tasks. Many existing CL strategies require extensive retraining or memory buffers to store past data, making them impractical for real-time industrial applications.

In this work, we propose a Multi-Expert Continual Anomaly Detection architecture (\methname). Our system dynamically assigns experts to object classes based on feature similarity, enabling efficient knowledge specialization.
This method eliminates the need for full model retraining and reduces computational overhead while ensuring that new object classes are integrated efficiently.  By distributing the learning tasks across specialized experts, our approach effectively mitigates catastrophic forgetting while maintaining high detection performance.

We demonstrate the effectiveness of our method on the widely used MVTec AD dataset, a standard benchmark for industrial anomaly detection. Our approach achieves an excellent anomaly detection performance across 15 diverse object categories while utilizing only 5 experts. 
Our system maintains a balanced distribution of computational resources while offering improved specialization for similar object categories.

The main contributions of our paper are as follows:
\begin{itemize}
    \item we present \methname, a multi-expert architecture for continual anomaly detection that enables the system to learn incrementally from sequentially introduced object classes while mitigating catastrophic forgetting through specialized knowledge distribution;
    \item we introduce a two-level memory management system that leverages Coreset selection with expert-specific replay buffers, optimizing storage efficiency while retaining critical information about previous classes; and
   \item We develop a similarity-driven expert assignment strategy that dynamically routes new classes to the most appropriate experts based on feature similarity measures, configurable thresholds, and capacity constraints, resulting in natural specialization patterns that improve both performance and knowledge retention.
    
\end{itemize}

\section{Related Work}
\label{sec:soa}


\subsection{Usupervised Anomaly Detection}
Anomaly detection (AD) methods aim to identify patterns deviating from normal data distributions and can be broadly categorized into reconstruction-based and feature-embedding-based techniques.
Reconstruction-based methods such as Variational Autoencoders (VAEs) \cite{kingma2013auto} and Generative Adversarial Networks (GANs) \cite{yan2021learning} learn to model normal data patterns and identify anomalies based on high reconstruction errors. 
Feature-embedding-based techniques, on the other hand, project data into a learned latent space where normal and anomalous samples can be better separated. Memory bank models \cite{roth2022towards,defard2021padim} and one-class SVMs \cite{deng2022anomaly} fall under this category. Recent works utilizing more advanced architectures like MoEAD \cite{meng2025moead} and ADMoE \cite{zhao2023admoe} have introduced the use of transformer-based mixture-of-experts for enhanced anomaly scoring, though they do not integrate continual learning.
Our work presents an evolution of PatchCore \cite{roth2022towards} for continual anomaly detection. Unlike prior works that use fixed models, our method incorporates specialized experts that can efficiently adapt to new classes while preserving knowledge of previously learned patterns.

\subsection{Continual Learning}
Continual learning (CL) focuses on sequential learning of multiple tasks without catastrophic forgetting. 
%
Regularization-based methods like Elastic Weight Consolidation (EWC) \cite{kirkpatrick2017overcoming,zenkeimproved,aljundi2018memory,nguyen2017variational} encode knowledge of previous tasks to constrain network weight updates. However, these methods often struggle with significant task dissimilarities.
%
%
Replay-based methods maintain a memory buffer containing representative samples from past tasks \cite{ER2018RDAVID,rebuffi2017icarl}.
Selection of informative samples is crucial, with approaches like Gradient-based Sample Selection (GSS) \cite{aljundi2019gradient}, Experience Replay with Maximally Interfered Retrieval (ER-MIR) \cite{aljundi2019mir} and Online Coreset Selection (OCS) \cite{yoon2021coreset} addressing this challenge.
%
%
Finally, Expert-based methods like Expert Gate \cite{aljundi2017expert} and Mixture-of-Experts (MoE) \cite{li2024theory} train multiple specialized models, each assigned to specific tasks \cite{le2024mixture,rypesc2024divide,yu2024evolve,hihn2021mixture}. TAME \cite{zhu2024tame} employs a task detection mechanism to activate the most relevant expert based on loss deviation. However, these approaches typically
focus on classification rather than unsupervised anomaly detection. 
%
%
Our approach combines replay-based principles with the multi-expert architecture for the specific challenges of continual anomaly detection, where the goal is to identify anomalies in sequentially introduced object classes rather than classify them.

\subsection{Continual Learning for Anomaly Detection (CLAD)}

Recent works have begun to address continual learning in anomaly detection contexts specifically, demonstrating how structured memory retention enhances knowledge transfer and reduces forgetting~\cite{crad}. A benchmark for continual anomaly detection with pixel-level evaluation has also been introduced~\cite{bugarin2024unveiling}. 
Proposed approaches include fitting of normal data distributions~\cite{li2022towards}, adaptive learning~\cite{dalle2022continual} and incremental learning~\cite{tang2025incremental} strategies, continual prompting and contrastive learning within a Vision Transformer~\cite{ucad}, and reverse distillation~\cite{yang2024reverse}. 
All these works demonstrated that simple replay strategies significantly outperform conventional approaches in dynamic environments~\cite{faber2024lifelong}.
Nevertheless, they often face trade-offs between adaptability, accuracy, and computational efficiency. Existing methods typically use fixed model architectures or lack explicit mechanisms for managing expertise across diverse object classes.
Our approach addresses these specific gaps by leveraging a multi-expert architecture with specialized memory management techniques, offering a more effective framework for handling the dynamic nature of industrial anomaly detection tasks compared to the methods discussed above.

\section{Multi-expert continual anomaly detection}

\begin{figure}[t]
  \centering
  \includegraphics[width=\linewidth]{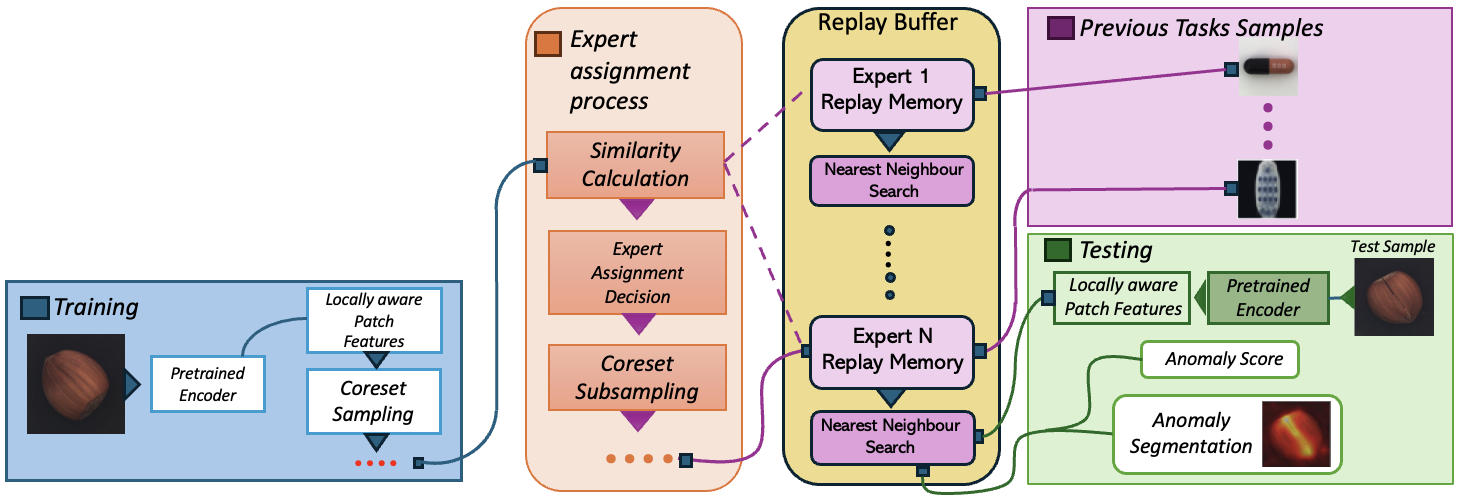}
  \caption{Overview of the \methname pipeline. The system processes object classes sequentially, extracting patch-level embeddings from normal samples, Coreset of 
  these embeddings are assigned to specialized experts based on cosine similarity. Each expert maintains a memory buffer with randomly selected samples and is updated only when assigned a new class. During inference, test samples are processed through the corresponding expert to generate anomaly scores at the image level.}
  \label{fig:pipelineme}
\end{figure}

Our framework for continual anomaly detection with a multi-expert architecture (\methname) is designed to incrementally learn and detect anomalies across multiple object classes while mitigating catastrophic forgetting. The system dynamically assigns experts to object classes based on feature similarity and employs efficient memory management for optimal performance. 
The overall pipeline of \methname is presented in Figure~\ref{fig:pipelineme}.

\subsection{Multi-Expert Architecture}
We use a fixed number of expert models $E = \{E_1, E_2, ..., E_N\}$, where each expert specializes in detecting anomalies for one or more object classes. 
Inspired by PatchCore~\cite{roth2022towards}, each expert operates as an independent memory bank of normal sample embeddings, while the feature extraction backbone is shared among all the experts. This architecture allows for specialized knowledge retention while enabling parallel processing during inference.
We limit the number of classes assigned to each expert, balancing specialization with computational efficiency. This design choice reduces interference between dissimilar classes while maintaining robust performance across the entire system.

\subsection{Patch embedding memory-bank}

Our framework relies on an efficient patch-level memory representation for anomaly detection. This approach is inspired by PatchCore’s mechanism \cite{roth2022towards}, but introduces novel mechanisms specifically designed for the continual learning context.
Given an input image $I \in \mathbb{R}^{H \times W \times 3}$, we extract discriminative patch-level features from intermediate layers of a pre-trained WideResNet50 [30] backbone. This produces a feature tensor $F \in \mathbb{R}^{h \times w \times d}$, where $h$ and $w$ represent spatial dimensions and $d$ is the feature dimension. The extracted patches are particularly effective because each patch embedding aggregates local descriptors from neighboring regions through the network's hierarchical receptive fields, capturing normal patterns at multiple scales.
During training, we build a memory bank of normal patch embeddings that serves as a reference for anomaly detection. For efficient resource utilization, we apply Coreset selection \cite{yoon2021coreset}  to reduce the full set of patch embeddings into a compact yet representative memory bank. This selection process identifies the most informative patches while discarding redundant information, allowing our system to maintain high detection performance with substantially lower memory requirements.
For anomaly detection at inference time, test image patches are compared against this memory bank using a nearest-neighbor approach. Patches that differ significantly from their nearest neighbors in the memory bank are flagged as potential anomalies, with the maximum patch-level anomaly score determining the image-level anomaly score.
Our multi-expert architecture extends this memory-bank approach by maintaining separate, specialized banks for different experts, each focusing on specific object classes based on feature similarity. This expert-specific memory organization is key to preventing catastrophic forgetting in the continual learning setting, as it allows the system to preserve critical information about previously seen classes while effectively incorporating new ones.

\subsection{Expert Assignment}
When a new class $c$ arrives, we determine which expert should handle it based on feature similarity. We assign the first class to the first expert. For subsequent classes, we compute the cosine similarity between the class's feature centroid and each expert's memory:
\begin{equation}
  \text{sim}(c, E_i) = \frac{\langle \mu_c, \mu_{E_i} \rangle}{||\mu_c|| \cdot ||\mu_{E_i}||}
\end{equation}
where $\mu_c$ is the centroid of embeddings of class $c$, $\mu_{E_i}$ is the centroid of expert $E_i$'s memory, and $\langle \cdot \rangle$ is the inner product operator.
The class is assigned to the expert with the highest similarity. To avoid assignments of classes to non-representative experts, we only assign the expert if the similarity exceeds a predefined threshold $\theta$. Otherwise, we assign the class to a new, unassigned, expert $E_{\emptyset}$ (if any).

\begin{equation}
  \mathcal{A}(c) = \begin{cases}
   \arg\max_{i} \text{sim}(c,E_i) & \text{if} \; \text{sim}(c,E_i) \geq \theta \\
   E_{\emptyset} & \text{otherwise}
  \end{cases}
\end{equation}



\subsection{Expert Training with Replay}
When a new class is assigned to an expert, only that expert is updated, leaving other experts unchanged. This targeted update strategy significantly reduces computational overhead compared to full model retraining. The expert is trained using both the new class embeddings and replay samples from previously assigned classes:
\begin{equation}
\mathcal{M}_{E_i} = S_c \cup \{r \mid r \in \mathcal{R}_{E_i,j}, j \in \mathcal{C}_{E_i} \setminus \{c\}\}
\end{equation}

where $\mathcal{M}_{E_i}$ is the updated memory for expert $E_i$, $S_c$ represents selected embeddings from the new class $c$, $\mathcal{R}_{E_i,j}$ represents the replay buffer samples for class $j$, and $\mathcal{C}_{E_i} \setminus \{c\}$ is the set of all classes previously assigned to expert $E_i$ except the current class. 



The replay mechanism is crucial for mitigating catastrophic forgetting. We maintain a fixed replay ratio to balance stability and plasticity.

\subsection{Anomaly Detection Inference}
During testing, for an image from class $c$, we extract patch embeddings and process them through the corresponding expert $E_i$ that was assigned to class $c$ during training.
For image-level anomaly detection, we compute the maximum patch-level anomaly score:
\begin{equation}
s_{\text{image}} = \max_{p \in P} s(p, \mathcal{M}_{E_i})
\end{equation}
where $P$ is the set of patch embeddings $p$ from the test image and $s(p, \mathcal{M}_{E_i})$ measures the distance between patch $p$ and the nearest neighbor in the expert's memory $\mathcal{M}_{E_i}$.

\vspace{1em}
\noindent Our framework achieves effective continual learning for anomaly detection by combining expert specialization, strategic class assignment, and efficient memory management. The experimental results demonstrate that this approach successfully mitigates catastrophic forgetting while maintaining high anomaly detection performance across sequentially introduced object classes.


\section{Experiments}

In this section, we evaluate our \methname framework on the MVTec AD dataset~\cite{bergmann2019mvtec}, a benchmark widely used for industrial anomaly detection tasks. We systematically analyze the impact of varying the number of experts on both anomaly detection performance and forgetting mitigation. Our experiments will provide detailed insights into how similarity-based expert assignment affects class-level performance and knowledge retention in continual learning scenarios. We also investigate expert utilization patterns and the effectiveness of our replay buffer mechanism in preserving knowledge of previously seen classes.

\subsection{Experimental Setup}

\subsubsection{Dataset}
MVTec-AD is a standard benchmark for industrial anomaly detection. The dataset consists of 15 different object categories 
with a total of 5,354 high-resolution images. Each category contains normal (defect-free) samples for training and both normal and anomalous samples for testing, with pixel-level ground truth annotations for anomalous regions.

\subsubsection{Evaluation Protocol and Metrics}
We follow a continual learning evaluation protocol where object classes are introduced sequentially. We report results with:
\begin{itemize}
\item \textbf{Image-level AUROC}: Area Under the Receiver Operating Characteristic curve for image-level anomaly detection
\item \textbf{Forgetting}: The decrease in performance on previously learned classes after incorporating new classes.
\item \textbf{Memory Usage}: The distribution of classes across experts and corresponding memory usage.
\end{itemize}

\subsubsection{Implementation Details}
We implemented our \methname framework using PyTorch with Amazon's PatchCore as the backbone architecture. For feature extraction, we used a WideResNet50 pre-trained on ImageNet, extracting features from layers 2 and 3. Input image size of $224 \times 224$ pixels and Patch size of  $32 \times 32$ pixels. We considered 400 a 400-samples-per-class memory budget and 2400 a 2400-samples-per-expert memory budget. 
We used random sampling to select the replayed samples, with a replay ratio of 0.2. The threshold for expert assignment is set to 0.9. 
All experiments were conducted on a single NVIDIA GPU with 16GB of memory. The training process was performed sequentially, with classes introduced in the same order across all experimental configurations to ensure fair comparison.

\subsection{Results and Analysis}
For our primary experiments, we systematically varied the number of experts from 1 to 8, while keeping all other hyperparameters consistent. This allowed us to isolate the impact of the expert allocation strategy on both performance and forgetting metrics.

\subsubsection{Class Performance Across Expert Configurations.} 
\begin{figure}[t]
\centering
\centering
\includegraphics[width=\linewidth]{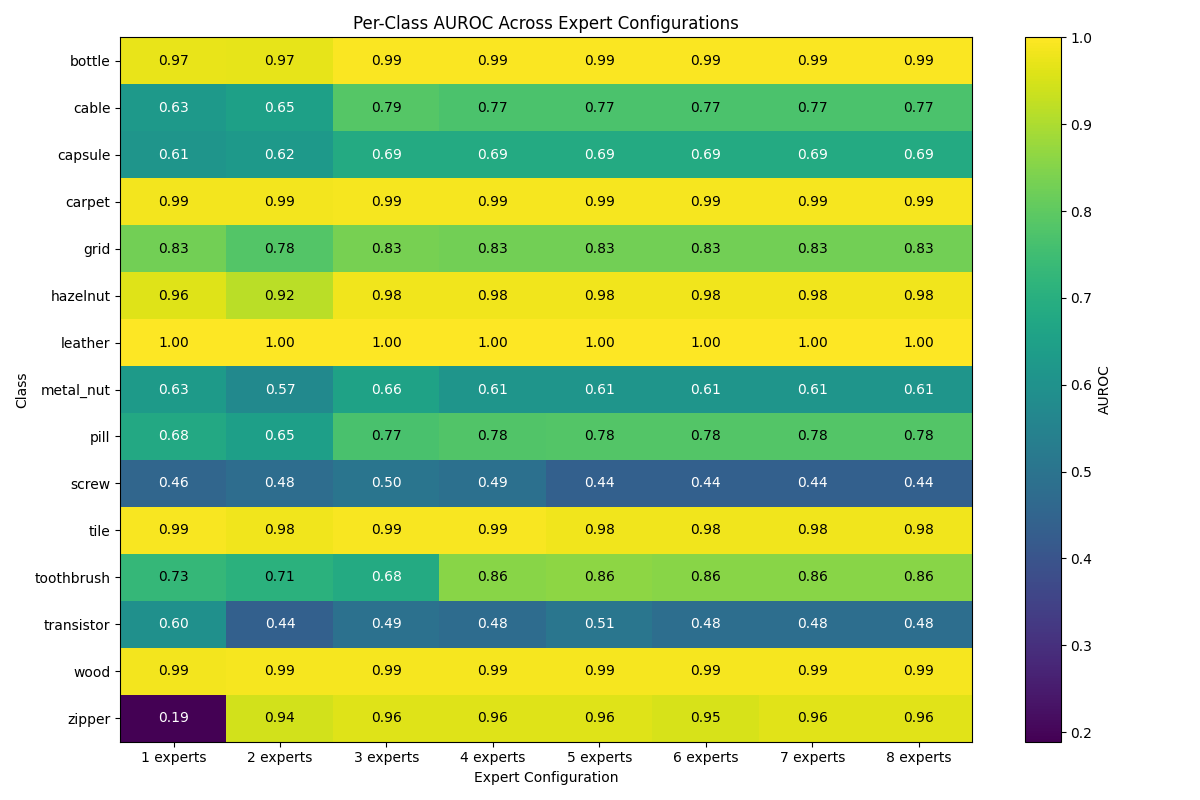}
\caption{Per-Class Auroc Across Expert Configurations}
\label{fig:class_auroc_heatmap}
\end{figure}

The heat map in Figure \ref{fig:class_auroc_heatmap} visualizes per-class AUROC across all expert configurations (1-8), revealing key patterns: certain classes maintain high performance (AUROC>0.95) regardless of expert count (e.g., leather, bottle), while others remain challenging (e.g., screw, transistor with AUROC<0.5). Most classes show significant improvement when moving from a single expert to multiple experts, with performance stabilizing after 3-5 experts. The heat map validates our similarity-based expert assignment approach, as evidenced by visible performance clusters in the 5-expert configuration, demonstrating our framework's ability to identify inherent class similarities for efficient knowledge distribution.

\subsubsection{Impact of Number of Experts}

\begin{figure}[t]
    \centering
    \begin{subfigure}{0.48\textwidth}
        \centering
        \includegraphics[width=\linewidth]{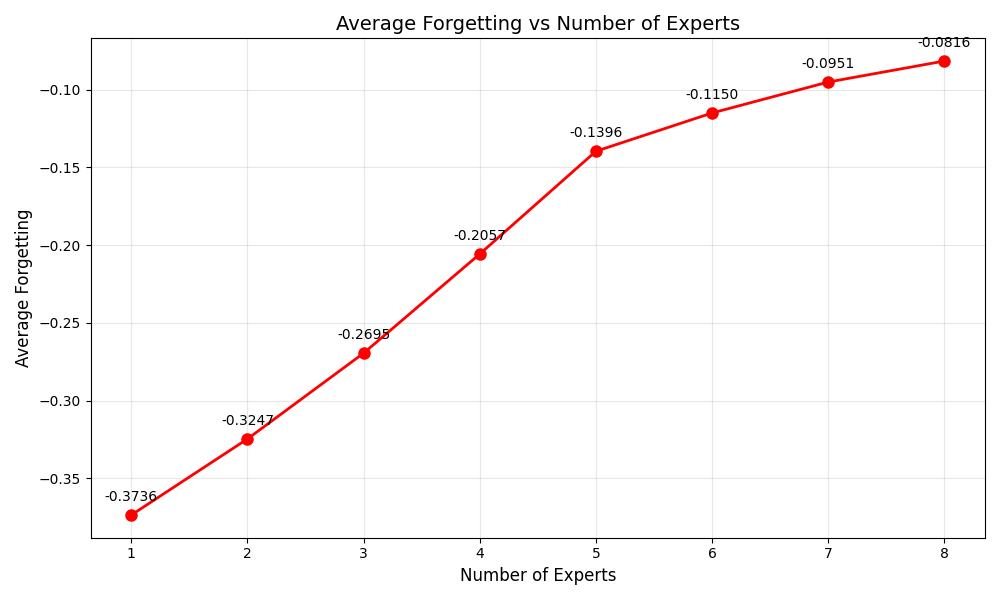}
        \caption{Average Forgetting across different numbers of experts}
        \label{fig:expert_forgetting}
    \end{subfigure}
    \hfill 
    \begin{subfigure}{0.48\textwidth}
        \centering
        \includegraphics[width=\linewidth]{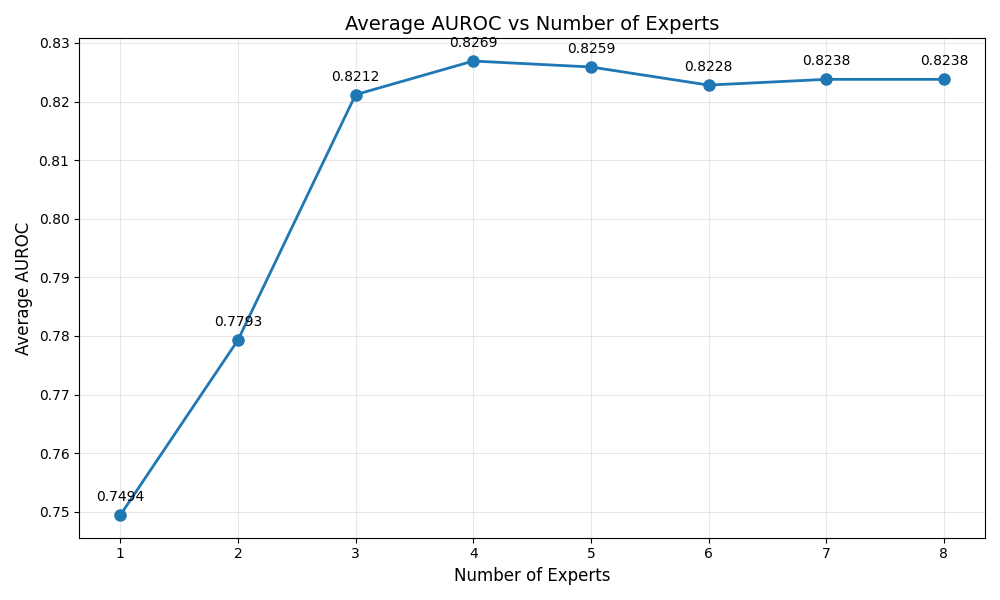}
        \caption{Average AUROC performance across different numbers of experts}
        \label{fig:Auroc_vs_experts}
    \end{subfigure}
    \caption{Impact of varying the number of experts (1-8) on (a) forgetting mitigation and (b) anomaly detection performance across all classes}
    \label{fig:expert_comparison}
\end{figure}

Figure\ref{fig:Auroc_vs_experts} illustrates the relationship between the number of experts and average AUROC across all 15 classes. The results demonstrate a clear pattern; single-expert configuration achieves 0.7494 AUROC. Performance increases substantially to 0.7793 with 2 experts. A significant jump to 0.8212 occurs with 3 experts. The peak performance of 0.8269 was reached with 4 experts. Performance stabilizes around 0.823-0.824 with 5-8 experts

This pattern indicates diminishing returns beyond 4 experts in terms of raw anomaly detection performance. However, examining the forgetting in Figure\ref{fig:expert_forgetting} reveals a different pattern; Forgetting decreases progressively as the number of experts increases. The single expert system exhibits severe forgetting (-0.3736) Substantial improvement occurs with 5 experts (-0.1396) Forgetting continues to decrease with 6-8 experts (-0.0816 with 8 experts).

These results demonstrate that while raw performance plateaus after 4-5 experts, additional experts continue to provide significant benefits for knowledge retention.

\subsubsection{Optimal Expert Configuration}
Based on the comprehensive analysis of both performance and forgetting metrics, we identify the 5-expert configuration as the optimal balance point for our \methname framework. This configuration achieves near-peak anomaly detection performance (0.8259 AUROC), while substantially reducing forgetting compared to configurations with fewer experts as well as limiting the memory consumption.






\subsubsection{Per-Expert and Per-Class Performance}
\begin{figure}[t]
    \centering
    \begin{subfigure}{0.48\textwidth}
        \centering
        \includegraphics[width=\linewidth]{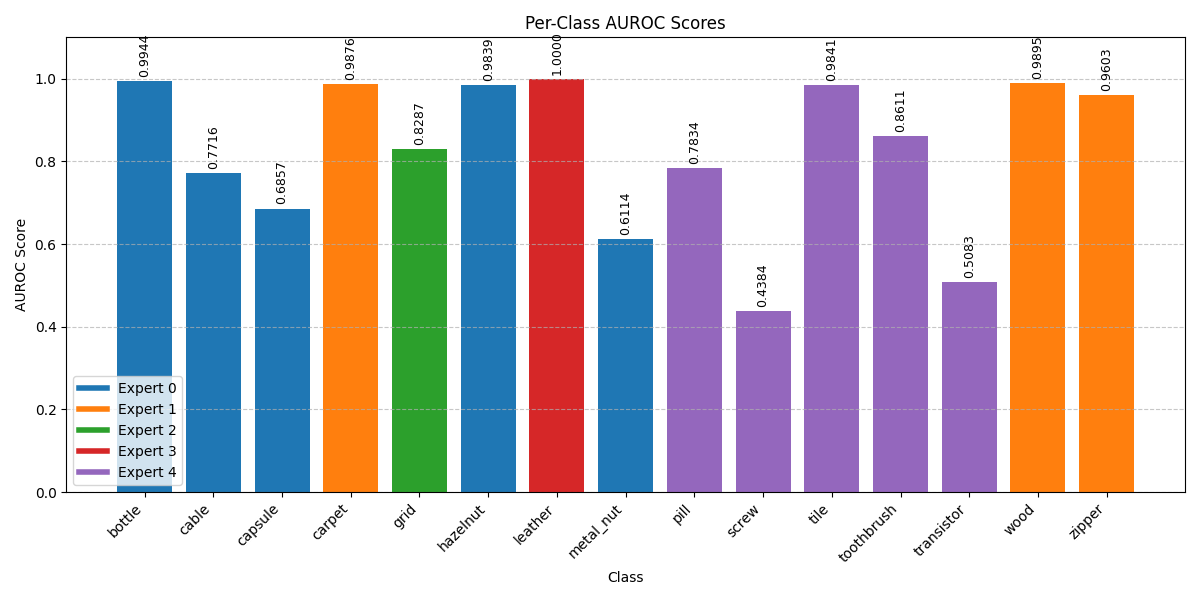}
        \caption{Per-class AUROC scores across all 15 classes}
        \label{fig:per_class_auroc}
    \end{subfigure}
    \hfill 
    \begin{subfigure}{0.48\textwidth}
        \centering
        \includegraphics[width=\linewidth]{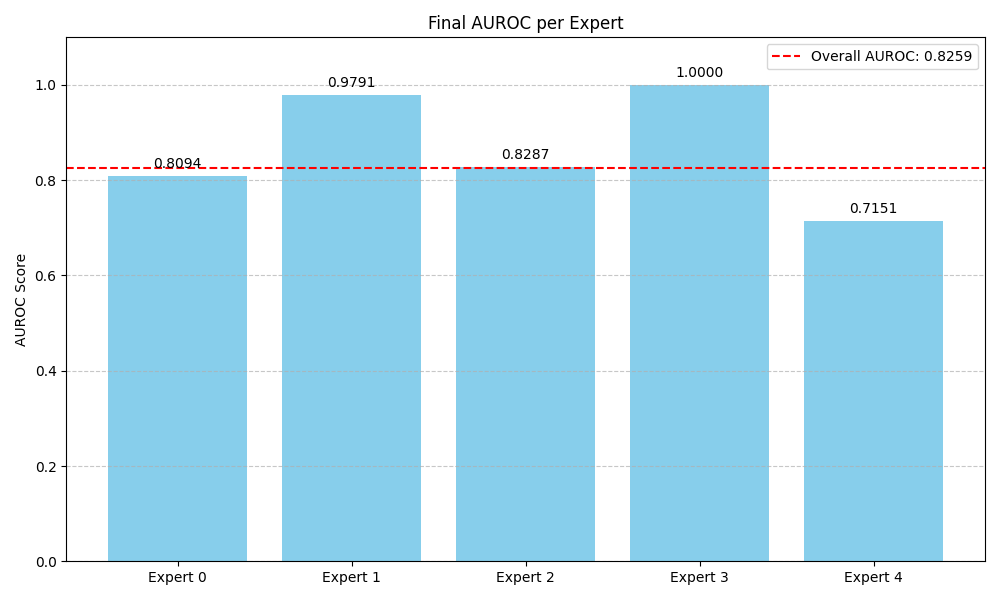}
        \caption{Final AUROC scores per expert}
        \label{fig:final_auroc}
    \end{subfigure}
    \caption{Final evaluation results showing (a) detailed per-class AUROC performance and (b) average AUROC scores achieved by each expert in the 5-expert configuration}
    \label{fig:final_metrics}
\end{figure}

Figure \ref{fig:final_auroc} presents the final AUROC scores for each expert in the 5-expert configuration. The substantial variation in performance across experts is notable: Experts 2 and 3 are assigned to a single class, providing results in line with the state of the art for that specific class. Expert 1 provides excellent performance on 3 texture-based classes. Experts 0 and 4 are assigned 5 classes each, with relatively high variability. 

The detailed per-class performance in Figure \ref{fig:per_class_auroc}  further illustrates this variation, with AUROC scores ranging from 0.4384 (screw) to 1.0000 (leather). This class-level analysis reveals that the low performance of some experts is mostly due to a few problematic classes.
These results demonstrate that our multi-expert architecture effectively handles a diverse range of object classes while maintaining strong overall performance.


\subsubsection{Memory Utilization Analysis.} Our approach maintained efficient memory management through controlled per-expert budgets. In the 5-expert configuration, memory utilization varied according to class assignments: Expert 0 reached 83.33\% utilization with its 5 assigned classes (bottle, cable, capsule, hazelnut, metal\_nut), Expert 1 utilized 50\% with 3 classes (carpet, wood, zipper), Experts 2 and 3 used only 16.67\% with single class assignments (grid and leather, respectively), while Expert 4 reached 83.33\% with 5 classes (pill, screw, tile, toothbrush, transistor). This distribution demonstrates how our similarity-based assignment naturally balances memory resources, with experts specializing in related classes utilizing similar memory proportions.

\subsubsection{Forgetting Analysis}
\begin{figure}[t]
    \centering
    \begin{subfigure}{0.48\textwidth}
        \centering
        \includegraphics[width=\linewidth]{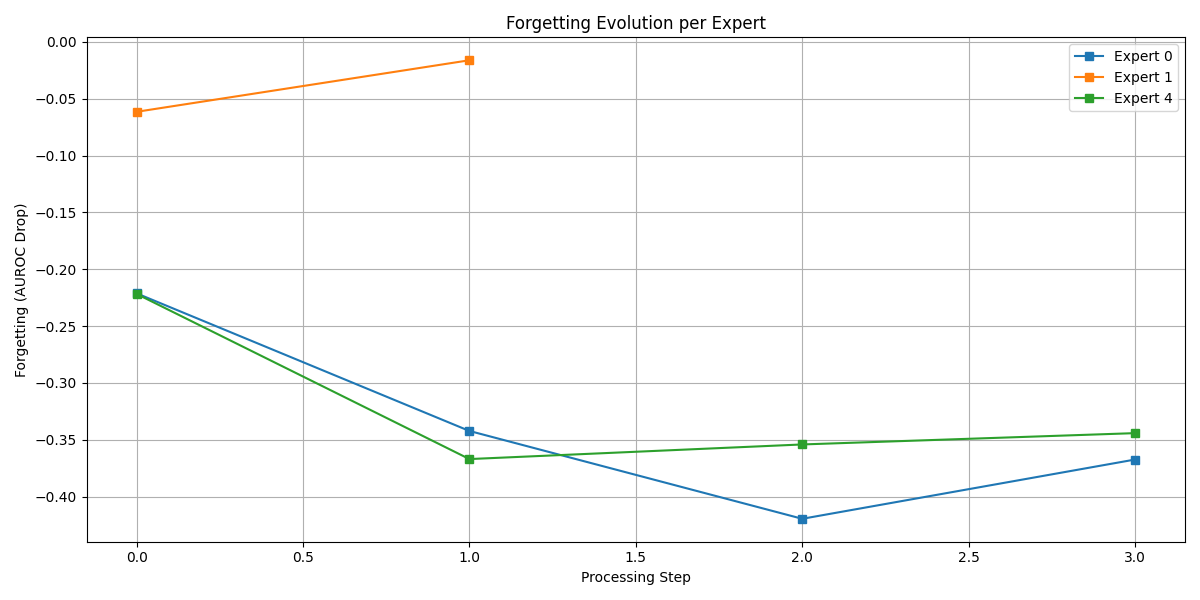}
        \caption{Forgetting evolution per expert}
        \label{fig:forgetting_evolution}
    \end{subfigure}
    \hfill 
    \begin{subfigure}{0.48\textwidth}
        \centering
        \includegraphics[width=\linewidth]{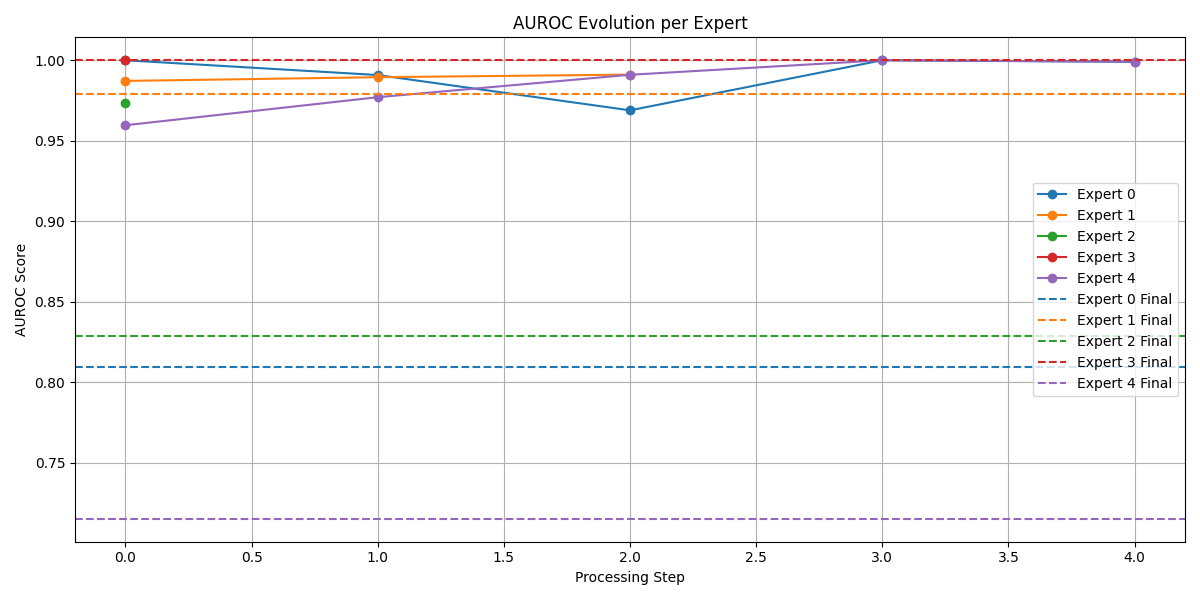}
        \caption{AUROC evolution per expert }
        \label{fig:auroc_evolution}
    \end{subfigure}
    \caption{Evolution of (a) forgetting and (b) AUROC metrics for each expert in the 5-expert configuration as classes are introduced sequentially}
    \label{fig:evolution_metrics}
\end{figure}

Figure \ref{fig:forgetting_evolution}  illustrates the evolution of forgetting for different experts as new classes are introduced. The data reveals distinct patterns:
Expert 1 (handling texture-based classes) maintains stable performance with minimal forgetting (-0.0144 in final evaluation)
Experts 0 and 4 (each handling 5 diverse classes) experience more significant forgetting (-0.3601 and -0.3304 respectively). 
Forgetting correlates strongly with both the number and diversity of classes assigned to each expert.


\section{Conclusion}
\noindent
Summarizing, our experiments with varying numbers of experts (1-8) revealed that similarity-based assignment naturally creates efficient expert allocation patterns. The system intelligently groups similar classes 
with the same expert rather than distributing them arbitrarily. With a fixed similarity threshold of 0.9, our approach effectively balances detection performance (0.8238 AUROC with 8 experts) and forgetting mitigation (-0.0816 with 8 experts).
Our \methname framework demonstrates that a 5-expert configuration provides the optimal balance between performance (0.8259 AUROC across 15 MVTec AD classes), forgetting mitigation, and computational efficiency. By adaptively assigning classes based on feature similarity and implementing targeted replay, our system effectively combats catastrophic forgetting without requiring full model retraining, making it well-suited for industrial environments with evolving product types.

\section*{Acknowledgements}
This study was carried out within the PNRR research activities of the consortium iNEST (Interconnected North-Est Innovation Ecosystem) funded by the European Union Next-GenerationEU (Piano Nazionale di Ripresa e Resilienza (PNRR) – Missione 4 Componente 2, Investimento 1.5 – D.D. 1058  23/06/2022, ECS\_00000043).

\par\vfill\par

\bibliographystyle{splncs04}
\bibliography{refs.bib}

\end{document}